\pdfoutput=1

\documentclass[11pt]{article}
\usepackage[table]{xcolor} 

\usepackage[final]{acl}

\usepackage{times}
\usepackage{latexsym}

\usepackage[T1]{fontenc}

\usepackage[utf8]{inputenc}

\usepackage{microtype}

\usepackage{inconsolata}

\usepackage{booktabs} 
\usepackage{graphicx}
\usepackage{amsfonts}
\usepackage[most]{tcolorbox}
\usepackage{float}
\usepackage{xspace}
\usepackage{bbding}
\usepackage{blindtext}
\usepackage{graphicx}
\usepackage{amsmath}
\usepackage{amssymb}
\usepackage{booktabs}
\usepackage{multirow}
\usepackage{amsmath, amssymb}
\usepackage{makecell}
\usepackage{graphicx, amsmath, amssymb, caption, subcaption, multirow, overpic, textpos}

\usepackage{arydshln}
\usepackage{wrapfig,lipsum,booktabs}
\definecolor{baselinecolor}{gray}{.9}

\newcommand\bluecolor[1]{\cellcolor{gray!40!blue!30}{#1}}

\newcommand\purplecolor[1]{\cellcolor{gray!20!purple!30}{#1}}

\tcbset{
  aibox/.style={
    width=474.18663pt,
    top=10pt,
    colback=white,
    colframe=black,
    colbacktitle=black,
    enhanced,
    center,
    attach boxed title to top left={yshift=-0.1in,xshift=0.15in},
    boxed title style={boxrule=0pt,colframe=white,},
  }
}
\newtcolorbox{AIbox}[2][]{aibox,title=#2,#1}

\title{LLaST: Improved End-to-end Speech Translation System Leveraged by Large Language Models}

\author{
Xi Chen$^{1}$, Songyang Zhang$^{2}$, Qibing Bai$^{1}$, Kai Chen$^{2}$, Satoshi Nakamura$^{1,3}$ \\
$^{1}$The Chinese University of Hong Kong, Shenzhen\\
$^{2}$Shanghai AI Laboratory\\
$^{3}$Nara Institute of Science and Technology, Japan\\
\texttt{\{xichen7@link,snakamura@\}.cuhk.edu.cn} \\}

\begin{document}
\maketitle
\begin{abstract}
We introduces \textbf{LLaST}, a framework for building high-performance \textbf{L}arge \textbf{La}nguage model based \textbf{S}peech-to-text \textbf{T}ranslation systems.
We address the limitations of end-to-end speech translation~(E2E ST) models by exploring model architecture design and optimization techniques tailored for LLMs. Our approach includes LLM-based speech translation architecture design, ASR-augmented training, multilingual data augmentation, and dual-LoRA optimization. Our approach demonstrates superior performance on the CoVoST-2 benchmark and showcases exceptional scaling capabilities powered by LLMs.
We believe this effective method will serve as a strong baseline for speech translation and provide insights for future
improvements of the LLM-based speech translation framework\footnote{We release the data, code and models in \href{https://github.com/openaudiolab/LLaST}{https://github.com/openaudiolab/LLaST}}.
\end{abstract}

\section{Introduction}

The speech-to-text translation (ST) task, which transcribes spoken language into written text in a different language, is pivotal for bridging communication barriers. This capability has a wide array of applications, including facilitating global communication, enabling automatic subtitling, and aiding in language learning.


Conventional ST systems are typically composed of two distinct components: an \textit{automatic speech recognition} (ASR) module that transcribes spoken speech into written text in the source language, and a \textit{machine translation} (MT) module that subsequently translates this text into the target language. These modules can be trained using paired ASR and text-to-text translation data, significantly enhancing the overall performance of ST systems. Despite their modular design, cascade systems are prone to error accumulation, where inaccuracies from the ASR stage are compounded in the MT phase, often leading to sub-optimal translations. Recently, the focus has shifted towards the development of end-to-end speech translation (E2E ST) models that bypass the need for separate automatic speech recognition (ASR) and machine translation (MT) modules by directly converting spoken input into text in the target language. Nonetheless, these approaches often necessitate extensive training datasets and are contingent upon sophisticated model architectures to achieve strong performance. 

Speech translation is intrinsically linked to natural language processing (NLP), as it involves the conversion of spoken language into written text in a target language, necessitating a deep understanding of both the source and target languages' linguistic structures and semantics.
The unprecedented capabilities that large language models (LLMs) have demonstrated across a variety of NLP tasks ~\cite{touvron2023llama, touvron2023llama2, achiam2023gpt} have opened up new possibilities to construct potent speech translation systems by leveraging these LLMs as a foundation. Recent research has seen some preliminary attempts  exploring this direction~\cite{chu2023qwen, wu2023decoder, huang2023speech}. Despite these advancements, the question remains on how to most effectively harness the vast potential of LLMs to develop a high-performance ST system in an efficient manner, without compromising on quality or scalability.


In this study, we focus on the exploration of best practices for constructing an effective speech translation system powered by Large Language Models (LLMs), which we term \textbf{LLaST}. The paper delves into the core aspects of the development process, specifically the \textit{model architecture design} and \textit{optimization techniques}.  Our exploration begins with the creation of a minimalist model architecture, examining the selection of key modules such as the speech encoder 
and LLMs. Subsequently, we investigate training strategies, including \textit{ASR-augmented training} and \textit{dual-LoRA optimization}. Moreover, to deepen our understanding of scaling laws in LLM-based ST, we also scrutinize the impact of 
model size variations. Through these concerted efforts, we aim to uncover insights that can significantly enhance the performance and training efficiency of LLaST.


Our contributions are listed as follows.

\noindent$\bullet$  We explore the LLMs-based speech translation method, including model architecture design, training strategies, and data recipe.

\noindent$\bullet$  Extensive evaluations demonstrate the superiority of our approach, surpassing the previous SOTA methods~\cite{barrault2023seamless} and achieving \textbf{45.1} BLEU on the \texttt{fr}$\rightarrow$\texttt{en} test set of CoVoST-2.

\noindent$\bullet$ We are dedicated to making all data recipes, training methodologies, and model weights associated with LLaST openly accessible to the community. By doing so, we foster transparency, collaboration, and advancement in the field of LLM-based speech translation technology. 

\section{Related Work}
\subsection{Cascaded Speech Translation}
Historically, the construction of speech translation systems has been approached in a cascading fashion, incorporating both an ASR and an MT subsystem~\cite{stentiford1988machine,ney1999speech, nakamura2006atr}. The procedure involves initially converting the input speech into text in the source language, which is subsequently translated into the target language. The primary objective of this line of research has been to mitigate error accumulation, including the use of multiple recognition outputs and the development of robust MT models~\cite{casacuberta2008recent,kumar2014some,sperber2017neural}. \citet{sperber2019self} introduces a self-attention mechanism to handle the lattice inputs, and \citet{zhang2019lattice} proposes a lattice transformer, equipped with a controllable lattice attention mechanism, to derive latent representations. \citet{lam2021cascaded} establishes a feedback cycle in which the downstream performance of the MT system serves as a signal to enhance the ASR system via self-training.

\subsection{End-to-End Speech Translation}
The development of end-to-end speech translation (E2E ST) models, which bypass the requirement for intermediary stages such as ASR outputs and lattices, has been a significant stride in mitigating error propagation. Research indicates that these E2E ST models demonstrate encouraging results and offer performance on par with cascaded models~\cite{sperber2019attention, ansari2020findings,bentivogli2021cascade,ye2021endtoend}. Moreover, these models present additional benefits such as lower latency and the potential to be applied to languages that lack a written form~\cite{berard2016listen}.

Data scarcity and the modeling burden are recognized as two significant obstacles impeding the performance of E2E ST~\cite{xu2023recent}. Firstly, the intrinsic complexity of speech translation, which integrates transcription and translation, presents a challenge in optimizing a single model to accomplish these cross-modal and cross-lingual tasks in one step. Secondly, ASR datasets are typically less extensive than MT datasets, and the extension to ST datasets further exacerbates this size discrepancy. To address this issue of data scarcity, researchers have employed strategies such as data augmentation~\cite{tsiamas2023segaugment,lam2022sample}, pre-training~\cite{wang2020curriculum,ao2022speecht5}, and knowledge distillation~\cite{liu2019end}, which leverage external datasets.

To alleviate the modeling burden, a variety of multi-task learning strategies have been investigated~\cite{zhang2018overview}. Originating from the multi-task encoder-decoder architecture~\cite{weiss2017sequence}, some researchers have chosen to split the decoder into two separate components~\cite{liu2020synchronous,anastasopoulos2018tied}: one dedicated to transcription and the other to translation. Parallel research efforts~\cite{liu2020bridging,cheng2023m} have similarly decoupled the encoder, with further work showing that a shared encoder can be independently partitioned~\cite{tang2021general,ye2022cross} to make better use of ASR data. In addition, non-autoregressive (NAR) modeling has been explored as a means to decrease latency~\cite{inaguma2021orthros,chuang2021investigating}.

Significantly, recent advancements have also delved into multi-tasking within the context of large-scale training, leading to impressive results on ST benchmarks. For instance, Whisper~\cite{radford2023robust} and SeamlessM4T~\cite{barrault2023seamless} have incorporated 680k and 470k hours of multilingual speech data in their training.

\begin{figure*}[!t]
  \centering
  \includegraphics[width=0.95\linewidth]{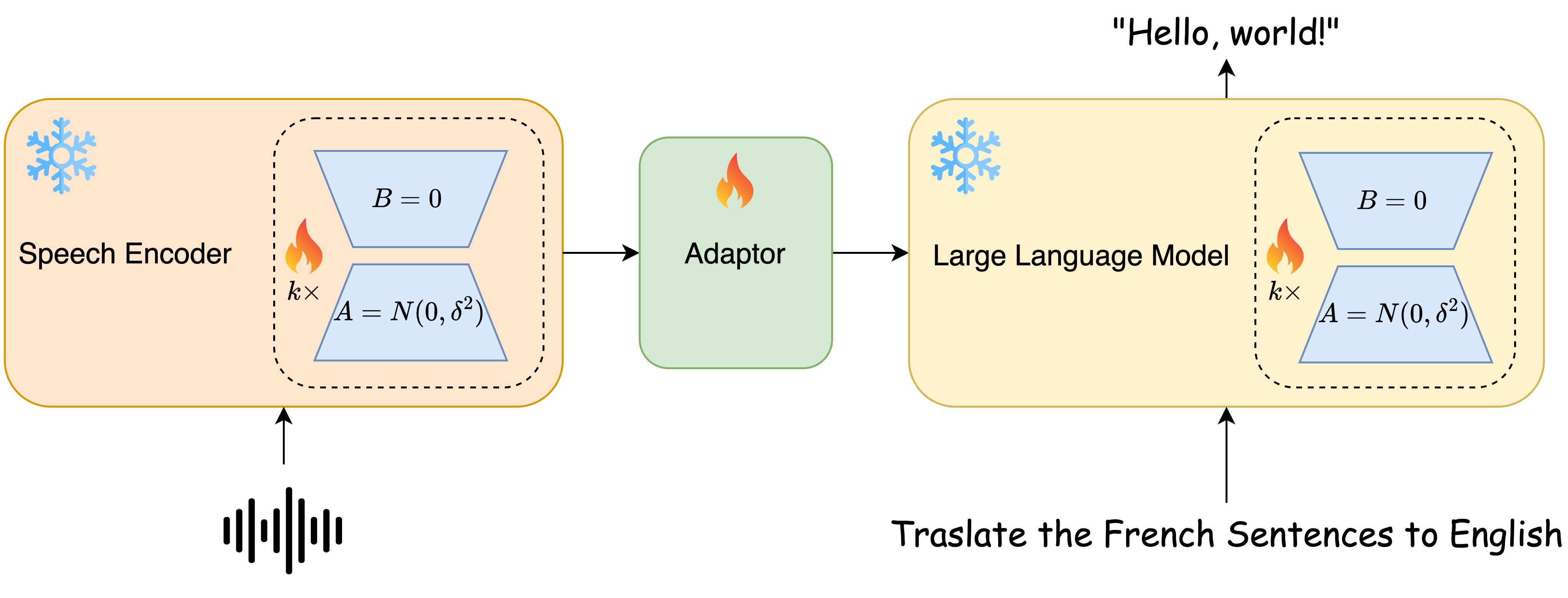}
  \caption{\textbf{Model Architecture of LLaST} We introduce \textit{dual-LoRA} in the optimization, and keep weights of the speech encoder and LLM frozen. We use a 3-layer MPLs for adaptor and fine-tune its parameters together with dual-LoRA.}
  \label{fig:structure}
\end{figure*}

\subsection{LLM-based Speech Translation}
Inspired by the robust linguistic capabilities of LLMs~\cite{brown2020language,touvron2023llama2}, recent initiatives have sought to harness the power of LLMs to address various speech tasks, aided mainly by instruction tuning. The prevailing method involves integrating an LLM (backend) with a speech encoder (frontend). Models like LauraGPT~\cite{chen2023lauragpt} and Qwen-audio~\cite{chu2023qwen} support a range of multi-modal speech tasks, demonstrating performance comparable to task-specific E2E ST models. VioLA~\cite{wang2023viola} employs a neural codec model~\cite{defossez2022high} to discretize the speech input while tuning the LLM. Similarly, AudioPaLM~\cite{rubenstein2023audiopalm} discretizes the speech input and achieves commendable results on CoVoST-2~\cite{wang2020covost}.

Salmonn~\cite{tang2023salmonn} employs two encoders as the frontend and uses LoRA~\cite{hu2021lora} for efficient fine-tuning. However, the extent of its performance improvement on ST remains largely unexplored. Some recent studies~\cite{wu2023decoder,zhang2023tuning} specifically target the ST task and delve into efficient tuning strategies, but their performance enhancements have been somewhat limited. In an industrial study focusing on translation between Chinese and English, \citet{huang2023speech} additionally incorporates the Chain-of-Thought (CoT) technique~\cite{wei2022chain}, enabling a step-by-step approach using LLMs.




\section{Method}

\begin{figure*}[!t] 
\begin{AIbox}{Example of Speech-text Prompt for LLaST}
{\bf \textcolor{black!40!green!60}{Speech Translation Prompt:}}  \hfill{\bf \textcolor{black!40!red!60}{Expected Output:}}\\
{
<audio>\textcolor{blue}{<AudioInputs>}</audio> Translate the French sentence to English.\hfill Hello world.\\
Transcripts of AudioInputs is "Bonjour le monde."
}
\tcbline
{\bf  \textcolor{black!40!green!60}{Automatic Speech Recognition Prompt:}} \hfill{\bf \textcolor{black!40!red!60}{Expected Output:}} \\
{
<audio>\textcolor{blue}{<AudioInputs>}</audio> Transcribe the French sentence to French.\hfill Bonjour le monde.\\
Transcripts of AudioInputs is "Bonjour le monde."
}
\end{AIbox} 
\caption{\textbf{An example for training data.}}
\label{fig:prompt}
\end{figure*}


This section presents our method in detail. We begin by introducing the problem setting of the speech-to-text translation task in Sec.~\ref{subsec:problemsetting}. Then, we explain the structure of  the proposed model in Sec.~\ref{subsec:structure}, followed by the description of the training and inference processes in Sec.~\ref{subsec:traininfer}.

\subsection{Problem Setting}
\label{subsec:problemsetting}
We now present the problem setting of speech translation. Given a speech translation dataset $\mathcal{D} = \{(\mathbf{S}, \mathbf{Y}_{src}, \mathbf{Y}_{tgt})\}$, the source language speech $\mathbf{S}$'s acoustic features~(e.g.,
mel-spectrogram) are denoted as $\mathbf{X}_s$, and we have: \[ \mathbf{X}_s=\mathcal{F}_{a}(\mathbf{S}),\quad \mathbf{X}_s = \{x_1, x_2, \textellipsis , x_T \} \]
where $\mathcal{F}_{a}$ is the acoustic feature extraction operation, and $T$ is the timesteps of the input features. $\mathbf{Y}_{src}$ and $\mathbf{Y}_{tgt}$ are the transcripts of $\mathbf{S}$ in the source and target languages, respectively.  The goal of speech translation is to generate the prediction text of target language $\mathbf{\hat{Y}}_{tgt}$ from the source speech $\mathbf{S}$. 

We can formulate the whole process as:
 \[ \mathbf{\hat{Y}}_{tgt} = \mathcal{F}(\mathbf{S}) \] and $\mathcal{F}$ represents the entire ST system.
 Performance of ST system is typically assessed by comparing the predicted output $\mathbf{\hat{Y}}_{tgt}$ with the ground truth $\mathbf{Y}_{tgt}$ using metrics like BLEU~\cite{papineni2002bleu}.

\subsection{Model Architecture}
\label{subsec:structure}
Our objective is to develop the LLaST model with a simple architecture, as depicted in Figure~\ref{fig:structure}. The design of LLaST comprises three key components: a speech encoder to process the input speech, an adaptor that projects these speech features into the compatible feature space for Large Language Models (LLMs), and finally, a decoder-only LLM for multi-modality decoding.


\paragraph{Speech Encoder}
Acoustic features $\mathbf{X}_s$ encapsulate a wealth of information, including speaker traits, emotions, prosody, background noise, and more. The role of the speech encoder is to disentangle these variabilities and generate robust linguistic representations, denoted as $\mathbf{Z}_s$. We define this process mathematically:
$$\mathbf{Z}_s = \mathcal{F}_{se}(\mathbf{X}_s)$$
where $\mathcal{F}_{se}$ represents the speech encoder function. Our work investigates various options for the speech encoder, with a focus on mHubert~\cite{hsu2021hubert, lee2021textless} and Whisper~\cite{radford2023robust}. For an in-depth analysis and discussion on the speech encoder selection, please refer to Sec.~\ref{exp:architecture}.


\paragraph{Adaptor}
The adaptor acts as a bridge between the speech encoder and the Large Language Model (LLM), consisting of a lightweight set of trainable parameters. Fine-tuning these parameters aligns speech features more effectively with the LLM's representation space. Its function is to project the extracted linguistic representations, $\mathbf{Z}_s$, into the embedding realm of the LLM, thus yielding $\mathbf{H}_s$:
$$\mathbf{H}_s = \mathcal{F}_{ada}(\mathbf{Z}_s)$$
This transformation process facilitates a smooth integration of speech data into the LLM's text-based context. We adopt a 3-layer multilayer perceptrons(MLPs) for adaptor.



\begin{table*}[!t]
\centering
\resizebox{0.8\textwidth}{!}{
\begin{tabular}{lllllll}
\toprule
\textbf{Model} & \textbf{Speech Encoder} & \textbf{Adaptor} & \textbf{LLM}\\
\midrule
LLaST-2B & Whisper-large-v2 & MLPs 
& TinyLlama-1.1B-Chat\\
LLaST-8B & Whisper-large-v2 & MLPs  & Llama2-7B-Chat\\
LLaST-14B & Whisper-large-v2 & MLPs  & Llama2-13B-Chat\\
\bottomrule
\end{tabular}}
\caption{
\textbf{Configurations of LLaST models}. We use Whisper(large-v2) and 3 layers MLPs for all LLaST models.
}
\label{tab:model_arch}
\end{table*}

\paragraph{Large Language Model}
Equipped with the projected speech feature $\mathbf{H}_s$, our objective is to utilize the Large Language Model (LLM) for generating the translated text of the original speech. To facilitate this, we construct a speech-text prompt input for the LLM. The text component of this prompt, denoted as $\mathbf{X}_q$, conveys the specific translation task instruction, such as "\texttt{Translate the French sentence into English}". Post-tokenization and embedding, $\mathbf{X}_q$ is transformed into the LLM's input representation, $\mathbf{H}_q$. 
Subsequently, the LLM generates translation predictions based on the concatenated speech-text features (for simplicity, we omit \texttt{bos} and \texttt{eos} tokens in the equation below):
$$\mathbf{\hat{Y}}_{tgt} = \mathcal{F}_{llm}([\mathbf{H}_s, \mathbf{H}_q])$$
This process allows the model to fuse speech and textual information effectively to produce translations.

In summary, the entire process can be expressed as:
$$\mathbf{\hat{Y}}_{tgt} = \mathcal{F}(\mathbf{S})=\mathcal{F}_{llm}([\mathcal{F}_{ada}(\mathcal{F}_{se}(\mathcal{F}_a(\mathbf{S}))), \mathbf{H}_q]) $$



\subsection{Training and Inference}
\label{subsec:traininfer}
This section delves into the optimization techniques employed in LLaST and elucidates its inference methodology.

\paragraph{Optimization with Dual-LoRA Fintuning}
To enhance training efficiency, we employ the LoRA~\cite{hu2021lora} tuning method for model optimization. This technique significantly reduces trainable parameters by introducing trainable rank decomposition matrices to each Transformer layer, while keeping the pre-trained weights frozen.

In LLaST, we introduce the \textit{dual-LoRA fine-tuning}, applying LoRA separately to both the speech encoder (S-LoRA) and the Large Language Model (L-LoRA). This approach ensures effective adaptation to speech translation tasks with minimal parameter updates. Specifically, we perform instruction-tuning on prediction tokens using the original auto-regressive training objective of LLM. For a target translation result $\mathbf{Y}_{tgt}$ of length $N$, its probability is calculated as:
$$P(\mathbf{Y}_{tgt}|\mathbf{X_s}, \mathbf{X}_q) = \prod_{i=0}^{N} P_{\theta}(y_i|\mathbf{X}_s, \mathbf{X}_q, \mathbf{Y}_{tgt,<i})$$
This strategy allows us to efficiently tune LLaST without extensive retraining, maintaining both computational efficiency and task-specific effectiveness.



\paragraph{Training with ASR-augmentation}
To enhance the performance of LLaST, we adopt the strategy from prior work \cite{barrault2023seamless, radford2023robust} to incorporate Automatic Speech Recognition (ASR) tasks for data augmentation during training. Given the structural similarity between ASR and ST tasks—both involve converting speech to text, we can simply modify the ASR prompt to match ST objectives, such as "\texttt{Transcribe the French sentence into English}". The examples of prompts are listed in Fig.~\ref{fig:prompt}. This ASR-augmentation significantly boosts the effectiveness of LLaST across various language pairs, as detailed in Sec.~\ref{exp:asr_augmentation}.


\paragraph{Inference Methodology}
During inference, we construct prompts in the same format as depicted in Fig.~\ref{fig:structure}. To generate translation text sequences $\hat{\mathbf{Y}}_{tgt}$, we employ a beam search algorithm with a beam size of 5.

\section{Experiments}

\begin{table*}[!t]
\centering
\resizebox{0.995\textwidth}{!}{
\begin{tabular}{l|c|llllll}
\toprule
\textbf{Model}& \textbf{Parms.} & \multicolumn{5}{c}{\texttt{X}$\rightarrow$ \texttt{English}} \\
& & French &  Japanese  &  German & Chinese  & Spanish & Italian\\
\hline
\multicolumn{8}{c}{\bluecolor{\textit{Baseline Models}}} \\
S2T\_Transformer~\cite{wang2020fairseq}&0.04B & 27.2 & N/A & 18.2 & N/A & 25.1 & N/A \\
SpeechLLaMA~\cite{wu2023decoder} & 7B& 25.2 & 19.9 & 27.1 & 12.3 & 27.9 & 25.9 \\
Whisper-small~\cite{radford2023robust} & 0.25B & 27.3 & 17.3 & 25.3 & 6.8 & 33.0 & 24.0\\
Whisper-large-v2~\cite{radford2023robust} & 1.6B &36.4 & 26.1 & 36.3 & 18.0 & 40.1 & 30.9 \\
Qwen-audio~\cite{chu2023qwen} & 8B& 38.5 & N/A & 33.9 & 15.7 & 39.7 & 36.0 \\
SeamlessM4T(medium)~\cite{barrault2023seamless} & 1.2B & 38.4 & 15.2 & 34.7 & 18.0 & 38.7 & 36.5  \\
SeamlessM4T(large-v2)~\cite{barrault2023seamless} &2.3B & 42.1 & 23.8 & 39.9 & 22.2 & 42.9 & 40.0 \\
\hline
\multicolumn{8}{c}{\purplecolor{\textit{Our Models}}} \\
\textbf{LLaST-2B} & 2B & 41.2 & 24.2 & 36.8 & 19.2 & 43.2 & 39.3\\
\textbf{LLaST-8B} & 8B & 44.1 &  24.4 &  40.8 &  23.3  &  45.3  &  42.1   \\
\textbf{LLaST-14B} & 14B & \textbf{45.1} &  \textbf{28.8}  & \textbf{41.2} & \textbf{24.8} & \textbf{46.1} & \textbf{43.0}\\
\bottomrule
\end{tabular}}
\caption{
\textbf{Performance comparison on CoVoST-2 \texttt{X}$\rightarrow$ \texttt{English} test set.} We use SacreBLEU scores as metrics for all experiments, the models are trained with multi-lingual data.
}
\label{tab:main}
\end{table*}

In this section, we conduct a series of experiments to validate the effectiveness of our method. We start by detailing experimental configurations in Sec.~\ref{sec:config}, followed by an overview of quantitative results in Sec.~\ref{sec:mainres}.
\subsection{Configurations}
\label{sec:config}
\paragraph{Datasets} Our speech translation models are trained and evaluated on CoVoST-2~\cite{wang2020covost}, a large-scale multilingual dataset that supports translations between English and 15 other languages, as well as from 21 languages into English. For monolingual experiments, we utilize six subsets with source languages translating to English, focusing on French-English for training and testing. In the multilingual setup, we employ \texttt{Fr}$\rightarrow$\texttt{En}, \texttt{Es}$\rightarrow$\texttt{En}, \texttt{De}$\rightarrow$\texttt{En}, \texttt{It}$\rightarrow$\texttt{En}, \texttt{Zh}$\rightarrow$\texttt{En}, and \texttt{Ja}$\rightarrow$\texttt{En} subsets 
and three English-to-X subsets: \texttt{En}$\rightarrow$\texttt{Zh}, \texttt{En}$\rightarrow$\texttt{Ja}, and \texttt{En}$\rightarrow$\texttt{De} 
. Audio samples are downsampled from 48kHz to 16kHz in all experiments.

\paragraph{Model Architecture} Tab.~\ref{tab:model_arch} presents the three LLaST model configurations. Each model utilizes a Whisper-large-v2 speech encoder, contributing approximately 1B parameters. The adaptor is a compact multilayer perceptron with three layers, ingesting 1280-dimensional inputs and adjusting its output dimensions to match those of the subsequent LLMs. Consequently, the overall parameter count is predominantly influenced by the LLM component. Hence, we denote our models as LLaST-2B, LLaST-8B, and LLaST-14B.



\paragraph{Hyperparameters} All models are optimized with AdamW, setting $\beta_1=0.9$ and $\beta_2=0.98$. A warmup-then-linear decay learning rate schedule is adopted, peaking at 0.0002. Training spans one epoch for each model. By default, the rank of S-LoRA (Whisper LoRA) is set to 128, while L-LoRA (LLM LoRA) rank is 512 unless specified otherwise. The LLaST-8B and LLaST-14B models are trained using 32 NVIDIA A100 GPUs, each with a batch size of 32, while the smaller LLaST-2B model is trained on a setup consisting of 8 A100 GPUs, maintaining the same batch size per GPU.
\subsection{Main Results}
\label{sec:mainres}

\paragraph{Comparisons with Other Models}
Tab.~\ref{tab:main} presents a comparison between our proposed LLaST models and previous methods, with SacreBLEU scores evaluated across six language pairs: \texttt{Fr}$\rightarrow$\texttt{En}, \texttt{Ja}$\rightarrow$\texttt{En}, \texttt{De}$\rightarrow$\texttt{En}, \texttt{Zh}$\rightarrow$\texttt{En}, \texttt{Es}$\rightarrow$\texttt{En}, and \texttt{It}$\rightarrow$\texttt{En}. Notably, LLaST-2B outperforms SeamlessM4T(medium) and demonstrates competitive performance against SeamlessM4T(large-v2). LLaST-8B significantly excels by improving upon the Qwen-audio model of similar scale with an impressive \textbf{5.6} BLEU point gain on the \texttt{Fr}$\rightarrow$\texttt{En} task. Furthermore, LLaST-14B achieves state-of-the-art (SOTA) results, attaining a BLEU score of \textbf{45.1} on CoVoST-2's \texttt{Fr}$\rightarrow$\texttt{En} subset, surpassing SeamlessM4T(large-v2) by \textbf{3.0} BLEU points. These results convincingly demonstrate the superiority of LLaST and highlight the promising potential of exploring LLMs for speech translation tasks.

\section{Ablation Analysis}

In this section, we delve into a meticulous ablation study and analysis of LLaST. We begin by examining the impact of model architecture in Sec.~\ref{exp:architecture}, followed by an exploration of optimization strategies in Sec.~\ref{exp:asr_augmentation}. Finally, we investigate the relationship between model scale and performance in Sec.~\ref{exp:model_scale}.

\subsection{Model Architecture Design}\label{exp:architecture}

\begin{table}[! t]
\centering
\resizebox{0.43\textwidth}{!}{
\begin{tabular}{lll}
\toprule
\textbf{Speech Encoder}& \textbf{LLM} & \textbf{BLEU}\\
\midrule
mHuBERT & TinyLlama  & 24.4 \\
Whisper-base & TinyLlama &  28.7 \\
\bottomrule
\end{tabular}}
\caption{\label{tab:audioencoders}
\textbf{Influence of different speech encoders.}  For speech encoder, mHuBERT-base(95M) and Whisper-base(74M) share the similar model size. We use TinyLlama-1.1B-Chat~\cite{zhang2024tinyllama} in this study. We report SacreBLEU scores on CoVoST-2 \texttt{fr}$\rightarrow$ \texttt{en} test set for all experiments. 
}
\end{table}
\paragraph{Choice of Speech Encoder}

We experiment with various speech encoder architectures, including mHuBERT~\cite{hsu2021hubert, lee2021textless} and Whisper~\cite{radford2023robust} model. For the mHuBERT, we adhere to the preprocessing approach from ~\cite{dong2023polyvoice, lee2021textless} to extract semantic units. 
For a fair comparison, we select the Whisper-base model, which is comparable in size to the mHuBERT model.
Performances reported in Tab.~\ref{tab:audioencoders} indicate that the Whisper model yields superior performance, demonstrating a \textbf{4.3} BLEU score improvement over mHuBERT. This improved performance can be attributed to the fact that Whisper has been trained on significantly more data, thus generating more representative linguistic features.

\begin{figure*}[!t]
  \centering
  \includegraphics[width=0.95\linewidth]{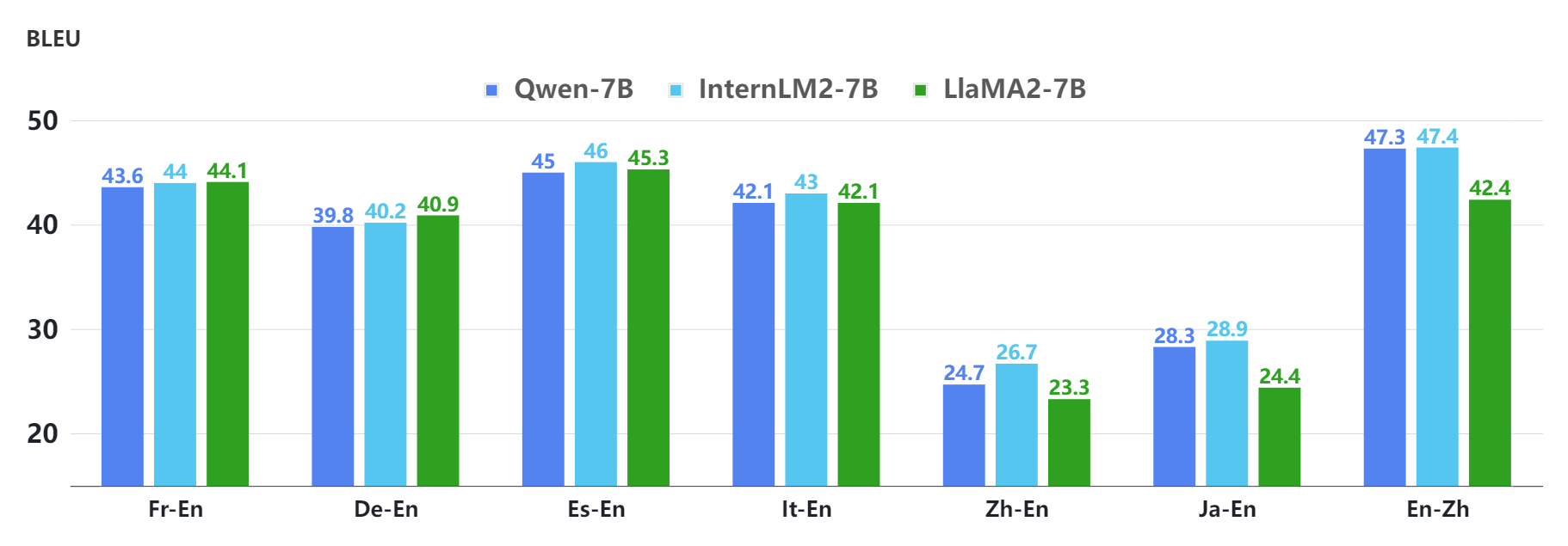}
  \caption{\textbf{Influence of different language models.} We use Whisper-large-v2 as speech encoder and report SacreBLEU scores on CoVoST-2 test set for all experiments.}
  \label{fig:diffllm}
\end{figure*}

\paragraph{Choice of Large Language Models}
We examine the impact of different large language models within LLaST to discern how variations in language modeling performance affect its speech translation capabilities. We present \texttt{X}$\rightarrow$\texttt{en} results in Figure~\ref{fig:diffllm}. Notably, Qwen achieves a score of 47.3 on the \texttt{en}$\rightarrow$\texttt{zh} test set, outperforming Llama2~\cite{touvron2023llama} by \textbf{4.9} BLEU points. Similarly, InternLM2\cite{cai2024internlm2} surpasses Llama2 by \textbf{5.0} BLEU points. These findings suggest that Chinese-oriented LLMs notably enhance performance on Chinese-related ST tasks, exemplified by \texttt{En}$\rightarrow$\texttt{Zh} and \texttt{Zh}$\rightarrow$\texttt{En}. 
The LLaST model, when coupled with Llama2, demonstrates exceptional performance particularly in the \texttt{Fr}$\rightarrow$\texttt{En} and \texttt{De}$\rightarrow$\texttt{En} language pairs. This intriguing observation underscores the potential of LLM-based ST approaches, as they allow for effortless integration of diverse LLM strengths tailored to specific languages or tasks.

\begin{figure}[!t]
  \centering
  \includegraphics[width=0.98\linewidth]{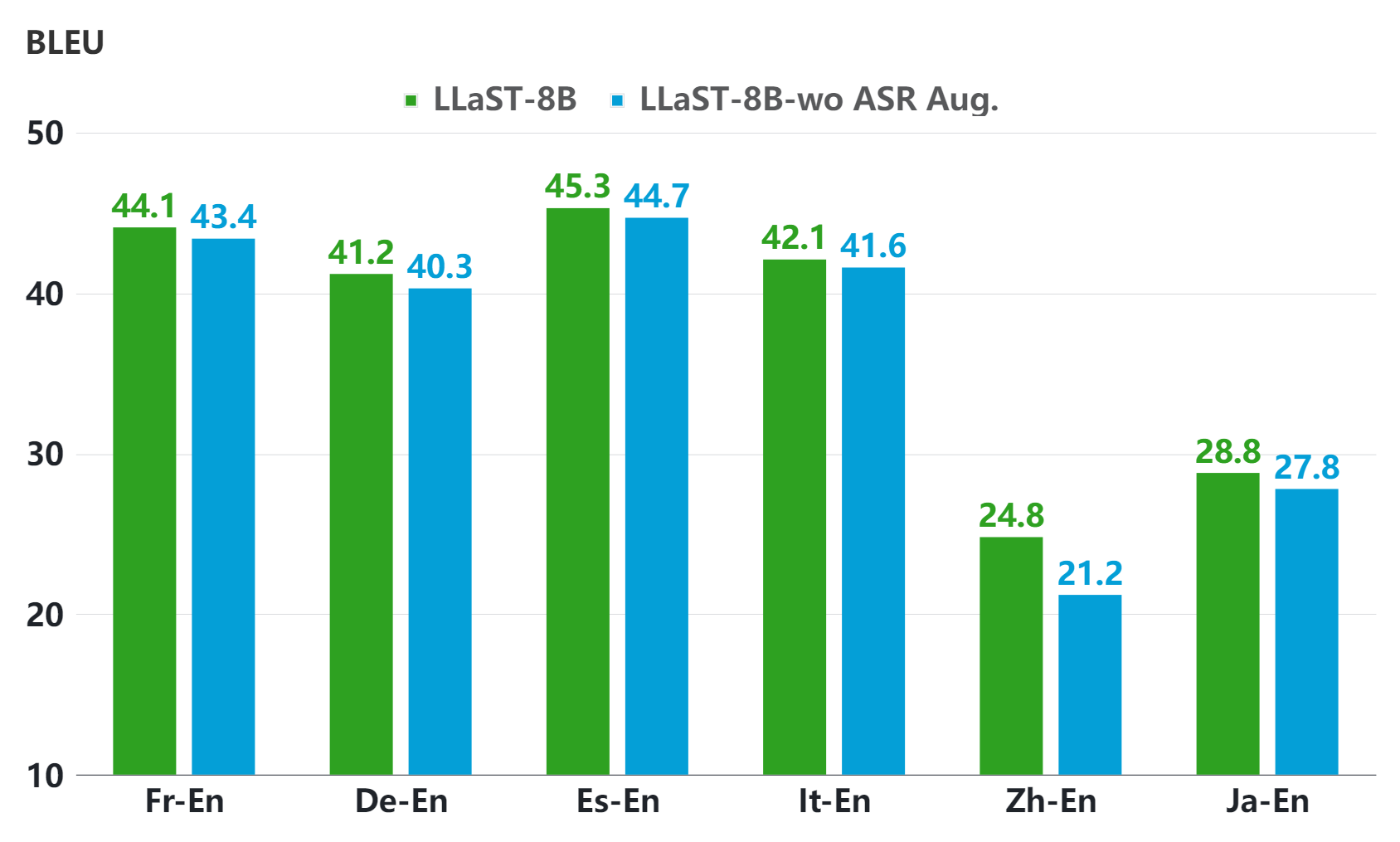}
  \caption{\textbf{Influence of different LLMs and ASR-augmentation.} We report SacreBLEU scores on CoVoST-2 test set for all experiments.}
  \label{fig:w_wo_asr}
\end{figure}




\subsection{Optimization}\label{exp:asr_augmentation}
\paragraph{Training with ASR Augmentation}
Automatic Speech Recognition (ASR) is a task akin to speech translation, as both involve converting speech into text. Prior research has leveraged ASR tasks as auxiliary objectives for ST training~\cite{zhang2018overview,ye2022cross,zhang2023rethinking}, or used models pre-trained on ASR data~\cite{wang2020fairseq}. In LLaST, we adopt this concept and incorporate ASR tasks to optimize LLaST performance. An example of the speech-text prompt structure can be found in Fig.~\ref{fig:prompt}, where ST and ASR samples are randomly mixed during training, with the focus remaining on the ST task at inference time.
The results presented in Fig.~\ref{fig:w_wo_asr} demonstrate the efficacy of ASR augmentation in optimizing LLaST. We observe across nearly all test sets that ASR augmentation improves ST performance, suggesting that leveraging ASR or multi-task training within LLM-based ST frameworks is a promising direction with significant potential for future work.

\paragraph{Multilingual Data Augmentation}
In our experiments, we explore both monolingual and multilingual settings. Specifically, for the monolingual setup, we employ the \texttt{Fr}$\rightarrow$\texttt{En} language pair. In the multilingual scenario, we introduce additional language pairs while maintaining the \texttt{Fr}$\rightarrow$\texttt{En} data identical to that in the monolingual experiment.

The results presented in Tab.~\ref{tab:w_wo_multilingualdata} reveal that incorporating other language pairs indeed benefits the \texttt{Fr}$\rightarrow$\texttt{En} translation task, with a \textbf{1.6} BLEU score improvement observed upon adding multilingual data augmentation. This finding aligns with similar phenomena reported in LLM research~\cite{team2023internlm,zeng2022glm}, where exposure to multilingual corpora has been shown to enhance the language modeling capabilities of these models.


\begin{table}
\centering
\resizebox{0.42\textwidth}{!}{
\begin{tabular}{lcl}
\toprule
\textbf{Speech Encode}& \textbf{Multi-Ling.}  & \textbf{BLEU}\\
\midrule
Whisper-large-v2 & \XSolidBrush & 42.5\\
Whisper-large-v2 & \Checkmark &  \textbf{44.1} \\
\bottomrule
\end{tabular}}
\caption{
\textbf{Study of training with multilingual data}. We use Llama2-7B-Chat for LLMs and report SacreBLEU scores on CoVoST-2 \texttt{fr}$\rightarrow$ \texttt{en} test set for all experiments.
}
\label{tab:w_wo_multilingualdata}
\end{table}

\paragraph{Dual-LoRA Optimization}
We investigate the impact of employing dual-LoRA for both speech encoders and large language models. In the ablation experiments, we utilize \textit{Whisper-large-v2} and \textit{Llama2-7B}. The results from scenarios without any LoRA, with LoRA applied only to Whisper, LoRA applied only to Llama2, and dual-LoRA are reported in Table~\ref{tab:duallora}.
From these outcomes, it is evident that even with a lightweight adaptor, leveraging a strong speech encoder and LLM can yield commendable performance. We also discover that applying single LoRA to either Whisper or Llama2 separately leads to substantial gains, improving scores from 40.5 to \textbf{41.3} and \textbf{43.6}, respectively. More notably, when dual-LoRA is used to jointly optimize both speech encoder and large language model, an additional improvement is achieved, culminating in a \textbf{44.1} BLEU score on test set.


\begin{table}[!t]
\centering
\resizebox{0.42\textwidth}{!}{
\begin{tabular}{cccl}
\toprule
\textbf{Adaptor} & \textbf{S-LoRA}& \textbf{L-LoRA} & \textbf{BLEU}\\
\midrule
\Checkmark & \XSolidBrush & \XSolidBrush & 40.5 \\
\Checkmark &\Checkmark & \XSolidBrush & 41.3 \\
\Checkmark &\XSolidBrush & \Checkmark  & 43.6 \\
\Checkmark &\Checkmark & \Checkmark  & \textbf{44.1} \\
\bottomrule
\end{tabular}}
\caption{
\textbf{Ablation study of dual-LoRA optimization strategy}. 
 S-LoRA means LoRA used in Whisper, and L-LoRA means the LoRA used in LLM. We use Whisper-large-v2 and Llama2-7B-Chat for speech encoder and LLMs, respectively. And we report SacreBLEU scores on CoVoST-2 \texttt{fr}$\rightarrow$ \texttt{en} test set for all experiments.}
\label{tab:duallora}
\end{table}

\subsection{Impact of Model Scale}\label{exp:model_scale}

\paragraph{Different Size of Speech Encoder}
We maintain a constant language model, \textbf{Llama2-7B}, and vary the size of Whisper models acting as speech encoders to examine the effect of encoder size on performance. The range of encoder sizes spans from 40M to 800M parameters. As shown in Table~\ref{tab:diff_wisper}, we observe that as the encoder size increases, the BLEU score of the model consistently improves; however, the rate of improvement diminishes with each incremental increase in size. The base encoder achieves a BLEU score of 37.0, while the large encoder attains a peak score of \textbf{44.1}. This considerable leap underscores the importance of scaling up speech encoders for better speech-to-text translation. However, future research should consider the trade-offs between model size, computational efficiency, and overall performance to strike the right balance for practical applications.


\paragraph{Different Size of LLMs}
We further investigate the impact of varying LLM sizes on speech translation performance. With the speech encoder consistently set as \textit{Whisper-large-v2}, we assess three different scale LLMs: TinyLlama-1B, Llama2-7B, and Llama2-13B. The outcomes are presented in Tab.~\ref{tab:main}.
Our findings reveal that there is a positive correlation between the size of the language model and the BLEU scores across all test sets. As the capacity of the LLM increases, so does the overall performance in terms of translation quality, indicating that larger models can capture more nuanced linguistic patterns and generate more accurate translations.


\paragraph{Comparison between the Encoder Scaling and Decoder Scaling} Given the Tab.~\ref{tab:diff_wisper} and Tab.~\ref{tab:main}, we observe some interesting phenomena. Despite Whisper-small+Llama2-7B-Chat and LLaST-2B demonstrating nearly equivalent performance on the fr->en subset, the former operates with approximately 7B parameters, whereas LLaST-2B functions with only about 2B parameters. This suggests that, in terms of parameter efficiency, scaling the encoder is a more effective strategy. It also indicates that, in these experiments, the encoder may play a more significant role. Meanwhile, the performance of the LLM-based system has yet to converge with respect to scale. To draw more comprehensive conclusions, we may need to continue scaling up the Whipster model and experiment with LLMs larger than 13B. For instance, in the domain of vision-language models, LLaVA~\cite{liu2023llava} and InternVL~\cite{chen2024internvl} demonstrate that achieving optimal performance with a larger vision encoder (6B) necessitates employing correspondingly larger LLMs, such as Yi-34B~\cite{ai2024yi}.

\begin{table}

\centering
\resizebox{0.45\textwidth}{!}{
\begin{tabular}{llll}
\toprule
 \textbf{Speech Encoder} & \textbf{Encoder Size}  & \textbf{BLEU}\\
\midrule
 Whisper-base  & $\sim$40 M &  37.0 \\
 Whisper-small  & $\sim$120 M &  41.2 \\
 Whisper-medium  & $\sim$390 M &  43.1 \\
 Whisper-large-v2  & $\sim$800 M & \textbf{44.1} \\
\bottomrule
\end{tabular}}
\caption{
\textbf{Ablation study of model size of Whisper model}. We use Llama2-7B-Chat for LLM and report SacreBLEU scores on CoVoST-2 \texttt{fr}$\rightarrow$ \texttt{en} test set.
}
\label{tab:diff_wisper}
\end{table}

\section{Limitation}
While our study has yielded significant findings, it is crucial to recognize the limitations that may impact the interpretation and broad applicability of our results. Although we delved into the architecture design and optimization strategies, our reliance on a relatively narrow data source and the use of short voice samples could potentially affect the generalizability of our outcomes. To address this, future research will expand to encompass a more diverse array of data. Moreover, due to the constraints of our current resources, we have not ventured into exploring larger language models or a broader range of language pairs in this study. 

 In speech translation, LLaST's use of LLMs raises concerns in actual application as teh following: (a)Probabilistic inaccuracy, mistranslations may occur due to nuances or dialects, impairing accuracy and cultural relevance.
(b)Data imbalances, insufficient representation in training data can lead to biased translations or reduced effectiveness for underrepresented groups.
(c)Deployment challenges, large model sizes and complexity may cause latency, high energy usage, and device compatibility issues.
(d)Harmful content generation, despite post-processing, risks persist; ongoing monitoring, filter refinement, and expert collaboration are needed.


\section{Conclusion}
We presents the development and analysis of LLaST, a novel speech translation model that harnesses LLM in this work. The study demonstrates that integrating well-tuned speech encoders like Whisper with different sizes of LLMs significantly improves speech-to-text translation performance. Through meticulous ablation studies, it is shown that applying dual LoRA optimization to both speech encoders and LLMs leads to substantial gains in BLEU scores. Additionally, experiments confirm that increasing the scale of either the speech encoder or the LLM positively impacts performance, though the rate of improvement decreases as size increases. Furthermore, incorporating ASR augmentation and multilingual training further enhances the model's performance on specific language pairs. Overall, LLaST underscores the potential of large language models for advancing speech translation tasks and offers valuable insights into their effective integration.
\section*{Ethical Considerations}
We use the public LLMs to build LLaST, the LLMs may produce unexpected outputs due to its size and probabilistic generation paradigm. For example, the generated responses may contain biases, discrimination, or other harmful content. Addtionally, we use ChatGPT and Grammarly to polish the writing.

\bibliography{custom}

\appendix

\end{document}